\documentclass[10pt,twocolumn,letterpaper]{article}

\usepackage{cvpr}              % To produce the CAMERA-READY version
\DeclareRobustCommand\onedot{\futurelet\@let@token\@onedot}
\def\eg{\emph{e.g. }}

\usepackage{url}
\usepackage{graphicx}
\renewcommand{\paragraph}[1]{\vspace{1mm}\noindent\textbf{#1}}
\usepackage{multirow, makecell}
\definecolor{baselinecolor}{gray}{.9}
\usepackage{xcolor, soul}
\sethlcolor{baselinecolor}
\usepackage{pifont}% http://ctan.org/pkg/pifont
\usepackage{colortbl}
\usepackage{xspace}
\usepackage{tabu}
\usepackage{amssymb}


\definecolor{cvprblue}{rgb}{0.21,0.49,0.74}
\usepackage[pagebackref,breaklinks,colorlinks,allcolors=cvprblue]{hyperref}

\usepackage{multirow}
\usepackage{epsfig}
\usepackage{graphicx}
\usepackage{amsmath}
\usepackage{amssymb}

\usepackage{booktabs}
\usepackage{epigraph}
\usepackage{enumitem}
\usepackage{float}
\usepackage{chngcntr}
\usepackage{soul}
\usepackage{caption}
\usepackage{subcaption}
\usepackage{setspace} 
\usepackage{zref-xr}
\usepackage{colortbl}
\definecolor{baselinecolor}{gray}{.8}
\DeclareRobustCommand\onedot{\futurelet\@let@token\@onedot}
\def\eg{\emph{e.g. }}

\usepackage{url}
\usepackage{graphicx}
\renewcommand{\paragraph}[1]{\vspace{1mm}\noindent\textbf{#1}}
\usepackage{multirow, makecell}
\definecolor{baselinecolor}{gray}{.9}
\usepackage{xcolor, soul}
\sethlcolor{baselinecolor}
\usepackage{pifont}% http://ctan.org/pkg/pifont
\usepackage{colortbl}
\usepackage{xspace}
\usepackage{tabu}
\usepackage{amssymb}

\usepackage{colortbl}
\usepackage{algorithm}
\usepackage{algorithmic}

\newcommand{\magenta}[1]{\textcolor{magenta}{{#1}}}

\newcommand{\figref}[1]{Fig.~\ref{#1}}
\newcommand{\tabref}[1]{Table~\ref{#1}}

\newcommand{\eqnref}[1]{Eqn.~(\ref{#1})}
\newcommand{\secref}[1]{Sec.~\ref{#1}}
\usepackage{xspace}
\usepackage{tabu}
\usepackage{amssymb}
\newlength\savewidth
\newcommand{\tablestyle}[2]{\setlength{\tabcolsep}{#1}\renewcommand{\arraystretch}{#2}\centering\footnotesize}

\usepackage{printlen}

\def\paperID{00000} % *** Enter the Paper ID here
\def\confName{CVPR}
\def\confYear{2025}

\title{VideoComp: Advancing Fine-Grained Compositional and Temporal Alignment in Video-Text Models}

\author{
Dahun Kim  \\
Google DeepMind\\
\and
AJ Piergiovanni \\
Google DeepMind\\
\and
Ganesh Mallya \\
Google DeepMind\\
\and
Anelia Angelova \\
Google DeepMind \\
}

\begin{document}
\maketitle

\begin{abstract}
We introduce VideoComp, a benchmark and learning framework for advancing video-text compositionality understanding, aimed at improving vision-language models (VLMs) in fine-grained temporal alignment. Unlike existing benchmarks focused on static image-text compositionality or isolated single-event videos, our benchmark targets alignment in continuous multi-event videos. Leveraging video-text datasets with temporally localized event captions (\eg ActivityNet-Captions, YouCook2), we construct two compositional benchmarks, ActivityNet-Comp and YouCook2-Comp. We create challenging negative samples with subtle temporal disruptions such as reordering, action word replacement, partial captioning, and combined disruptions. These benchmarks comprehensively test models’ compositional sensitivity across extended, cohesive video-text sequences. To improve model performance, we propose a hierarchical pairwise preference loss that strengthens alignment with temporally accurate pairs and gradually penalizes increasingly disrupted ones, encouraging fine-grained compositional learning. To mitigate the limited availability of densely annotated video data, we introduce a pretraining strategy that concatenates short video-caption pairs to simulate multi-event sequences. We evaluate video-text foundational models and large multimodal models (LMMs) on our benchmark, identifying both strengths and areas for improvement in compositionality. Overall, our work provides a comprehensive framework for evaluating and enhancing model capabilities in achieving fine-grained, temporally coherent video-text alignment. 
Dataset available at: \href{https://github.com/google-deepmind/video_comp}{\magenta{https://github.com/google-deepmind/video\_comp}}.
\vspace{-2mm}
\end{abstract}    
\begin{figure}[t]
   \centering
  \includegraphics[width=\linewidth]{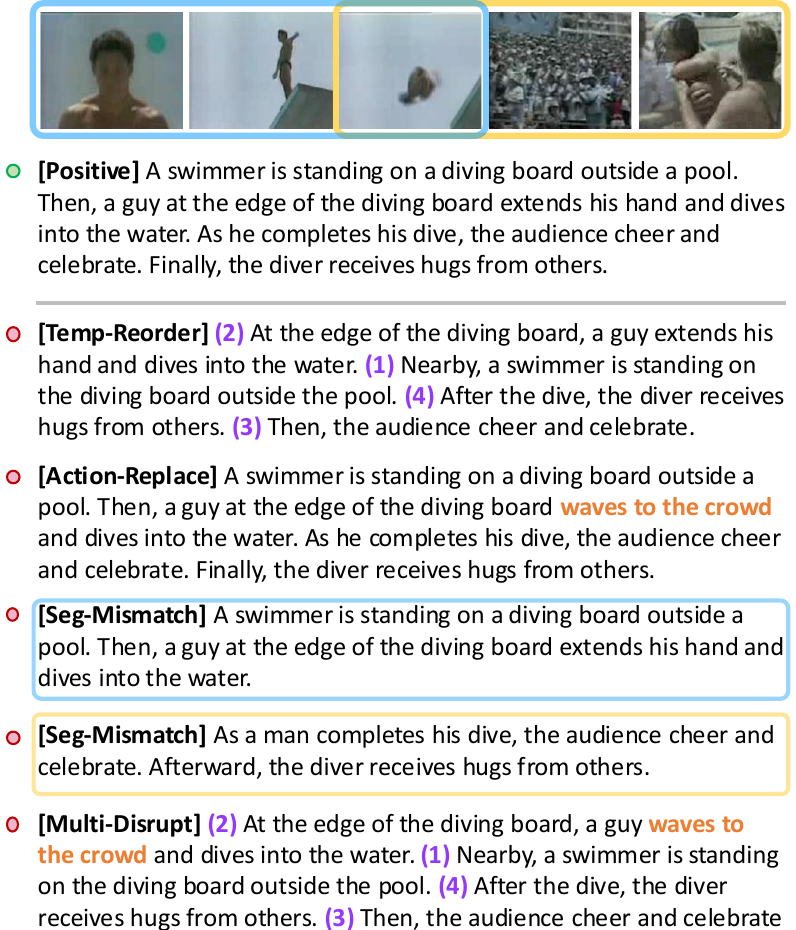}
   \vspace{-6mm}
   \caption{Illustration of our video-text compositionality benchmark, introducing challenging disruptions for fine-grained alignment. 
   Starting from a video and its positive text description (top row), we apply subtle disruptions including temporal reordering (purple), action word replacement (orange), and segment-level mismatch (yellow text matched with blue video crop) for evaluation, along with combined disruptions used during training. These perturbations test the model’s ability to distinguish coherent video-text pairs from disrupted ones.
   }
   \label{fig:teaser}
   \vspace{-5mm}
\end{figure}

\section{Introduction}
\label{sec:intro}

Compositionality, the capacity to understand and integrate attributes, relations, and entity states, is essential for vision-language models (VLMs) to achieve a comprehensive understanding of complex scenes. While traditional benchmarks largely focus on compositionality in static image-text pairs, extending this to video data requires models to capture both temporal and structural relationships within event sequences. 
In video compositionality, models must align temporally ordered and semantically rich video segments with text, discerning evolving relationships among entities within and across scenes. Although video-text understanding has gained attention, few benchmarks specifically address the dynamic, fine-grained temporal compositionality needed to process multi-event coherence in video. 

While compositional reasoning has been explored in image-text models, it has primarily focused on static scenes using contrastive losses and negative samples from text augmentations or alternative image sampling \citep{yuksekgonul2022and, doveh2023teaching}. These methods lack the understanding of temporal complexity required for video contexts. 
Recently, benchmarks such as PerceptionTest~\cite{patraucean2023perception}, VITATECS~\cite{VITATECS}, ViLMA~\cite{ViLMA}, and ICSVR \citep{madasu2023icsvr} have extended compositional evaluations to video, introducing counterfactual and concept-based tests. Yet, these benchmarks largely focus on isolated, single-event scenarios, which do not fully capture the continuous, multi-event coherence needed for real-world video understanding. 
Addressing this gap, our benchmark introduces multi-event video-text sequences with nuanced disruptions such as video segment cropping, action word replacements, and subtle temporal reordering to evaluate models’ ability to capture compositional alignment across cohesive video-text sequences.

In this work, we study video-text compositionality and temporal understanding by introducing a benchmark and learning framework that extends VLM capabilities. We base our benchmark on dense video captioning datasets like ActivityNet-Captions~\cite{krishna2017dense} and YouCook2~\cite{youcook}, which offer temporally localized captions for multiple events within each video. 
Using these datasets, we create structured modifications to evaluate video-text compositionality. For each video, we first obtain a positive sample by arranging event captions chronologically into a single, cohesive paragraph. To study compositional sensitivity, we then generate challenging negative samples with subtle disruptions, including temporal reordering (shuffling captions out of sequence), action word replacements (altering key verbs to shift meaning), segment-based misalignments (pairing partial video segments with mismatched captions), and combined disruptions. These nuanced modifications present models with difficult distinctions between coherent and disrupted sequences, particularly given the longer, interdependent annotations in our benchmark.

To improve model performance on video-text compositionality, we introduce a hierarchical pairwise preference loss alongside the standard InfoNCE objective. This preference loss guides models to prioritize similarity for temporally accurate pairs, while progressively reducing similarity scores for increasingly disrupted pairs. This approach enhances alignment with accurately composed samples and strengthens the model’s ability to recognize compositionality at varying levels of granularity.

One challenge in training for video-text compositionality is the limited availability of densely captioned, multi-event video datasets. To address this, we implement a pretraining strategy that leverages video-short text datasets. By concatenating short video clips and captions to simulate multi-event sequences, this pretraining method applies the preference loss to these pseudo-long-form pairs, enabling the model to develop a more robust understanding of temporal and compositional structure, even with limited dense-captioned data. This comprehensive approach allows models to better capture fine-grained, temporally coherent video-text alignment.

Finally, we evaluate large multimodal models (LMMs) on our benchmark. This evaluation provides insights into the current capabilities of LMMs in handling video-text compositionality, particularly in distinguishing temporally accurate alignments from corrupted ones.
The contributions of this work are as follows: 

\begin{itemize} \item We introduce a video-text compositionality benchmark featuring complex negative samples that disrupt the temporal alignment, providing a rigorous test for VLMs to capture fine-grained compositional structures.

\item We propose a hierarchical pairwise preference loss that adjusts video-text similarity scores based on levels of disruption, enhancing fine-grained alignment between video and text.

\item To facilitate video compositionality learning and address the scarcity of densely captioned video data, we explore a pretraining strategy that concatenates short video-caption pairs to simulate long-text data, promoting stronger temporal compositionality learning.

\item Our evaluation of large multimodal models reveals specific areas for improvement in compositional alignment.

\end{itemize}

\section{Related Work}
\label{sec:related}

\subsection{Video-Text Alignment and Understanding}

Video-text alignment models~\cite{lei2021less, bain2021frozen, wang2022omnivl, luo2022clip4clip, wang2023all, chen2022litevl, fu2021violet} are fundamental for tasks like retrieval, captioning, and question answering. Dual-encoder models~\cite{xu2021videoclip,luo2022clip4clip,all_in_one}, which adapt the CLIP~\cite{radford2021clip} framework for video by incorporating temporal elements, are efficient for large datasets and perform well in video-text retrieval tasks. However, they often lack the fine-grained temporal and compositional understanding needed for complex, multi-event sequences. 
While cross-attention models~\cite{videococa,li2022blip,ma2022x} can capture more detailed interactions between video frames and text tokens, they typically involve higher computational costs. 
Commonly used datasets, such as MSR-VTT~\cite{xu2016msr}, MSVD~\cite{msvd}, LSMDC~\cite{lsmdc}, WebVid~\cite{webvid}, InternVid~\cite{viclip} and VideoCC3M~\cite{nagrani2022learning}, support high-level alignment tasks but generally focus on short, single-event captions, limiting their effectiveness in evaluating temporal coherence within extended, multi-event contexts.
Several works \cite{zhang2023multi,ma2024ea,gwilliam2023video} have addressed multi-event retrieval by linking video events with specific text descriptions, focusing mainly on event identification rather than fine-grained temporal coherence within events.
Our benchmark extends these efforts, using dense captioning datasets with multi-event narratives to evaluate compositional coherence and temporal alignment in video-text models. By introducing targeted disruptions, we provide a comprehensive assessment of models' capabilities in maintaining compositional and detailed temporal understanding.
Our benchmark diverges from these efforts, using dense captioning datasets with multi-event narratives to evaluate compositional coherence and temporal alignment in video-text models. By introducing targeted disruptions, we provide a comprehensive evaluation of models' capabilities in maintaining compositional and fine-grained temporal understanding.

\subsection{Compositionality in Vision-Language Models}
It is essential for vision-language models to develop a structured understanding of language and its alignment with visual content, recognizing entities and capturing their fine-grained relationships.  In the image-text domain, compositional reasoning has traditionally been introduced to support this goal \citep{yuksekgonul2022and, kamath2023s,ma2023crepe,wang2023equivariant,zhang2023contrasting,kamath2023text, zhao2022vl, thrush2022winoground, singh2023coarse, parcalabescu2021valse}, focusing on attributes, relations, and object states within images. Previous approaches \citep{yuksekgonul2022and, doveh2023teaching} often use contrastive loss with negative samples generated through text augmentation or alternative image sampling to build compositional understanding. However, these methods primarily address static scenes, limiting their effectiveness in temporally dynamic video contexts. 

Later benchmarks, like PerceptionTest~\cite{patraucean2023perception}, introduce compositional tasks across multiple modalities but maintain video alignment as a secondary focus. Some video-specific benchmarks emphasize counterfactual and concept-based evaluations. VITATECS~\cite{VITATECS} tests models with counterfactual examples focused on temporal concepts within single-event video clips, while ViLMA~\cite{ViLMA} isolates specific temporal changes, such as shifts in actions or outcomes, within brief video segments. ICSVR \citep{madasu2023icsvr} examines the impact of compositional reasoning on video retrieval performance. Although these benchmarks advance compositionality in video, they largely assess isolated, single-event scenarios, leaving a gap in evaluating continuous, multi-event coherence. By contrast, our benchmark introduces complex, multi-event video-text sequences with nuanced disruptions such as video-text segment cropping, action word replacement, and subtle temporal reordering, challenging models to sustain compositional alignment across longer, more cohesive narratives.

\subsection{Temporal Compositionality Benchmarks for Video-Text Models}
Temporal compositionality benchmarks evaluate models’ understanding of time-based ordering and event sequencing. Benchmarks like Test of Time~\cite{Bagad2023Test} assess basic temporal order through binary before/after arrangements, while PerceptionTest~\cite{patraucean2023perception} examines temporal alignment across multiple modalities, though it mainly addresses broader multimodal capabilities. SEED-Bench~\cite{li2023seed}, VideoBench~\cite{ning2023video}, and MVBench~\cite{li2024mvbench} expand temporal assessment by incorporating multi-step sequences within procedural or spatial contexts but focus on shorter video segments. TempCompass~\cite{TempCompass} introduces more specific temporal alignment tasks, evaluating action sequencing, speed, and directional changes by reordering or reversing video segments. However, these benchmarks largely emphasize broad event sequencing rather than continuous, multi-event coherence seen in real-world scenarios.
Our benchmark moves beyond simple event ordering by presenting fine-grained temporal disruptions, such as overlapping segment cropping and gradual misalignment, across extended video sequences. This focus challenges models to retain coherence in multi-event contexts, offering a comprehensive assessment of temporal compositionality that better aligns with complex, real-world narratives.

\section{Method}
\label{sec:method}

We introduce a comprehensive framework to enhance compositional understanding in video-text models. The method involves constructing a benchmark dataset with compositional disruptions, designing compositionality-aware training objectives, implementing a pretraining strategy with short-form video-text data, and defining evaluation metrics that assess both broad video-text alignment and fine-grained compositional coherence.

\begin{figure}[t]
   \centering
   \includegraphics[width=0.99\linewidth]{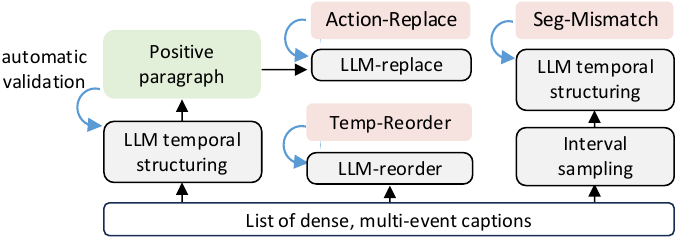}
   \vspace{-2mm}
   \caption{
    Overview of the dataset construction process. We start from a list of dense, multi-event captions of each video to obtain the positive text, then generate various negative texts with compositional disruptions.}
   \label{fig:overview}
   \vspace{-5mm}
\end{figure}

\subsection{Dataset Construction}
\label{sec:data}

Our video-text compositionality benchmark, ActivityNet-Comp and YouCook2-Comp, is designed to evaluate models’ abilities to capture fine-grained, temporally coherent alignment in multi-event video sequences. While standard video-text datasets like MSR-VTT~\cite{xu2016msrvtt} and ActivityNet~\cite{heilbron2015activitynet} often pair videos with brief, summarized captions, these short captions limit models’ understanding of the detailed temporal structure essential for complex event sequences. 
To address this gap, our benchmark extends the dense video captioning datasets ActivityNet-Captions~\cite{krishna2017dense} and YouCook2~\cite{youcook} by introducing targeted compositional disruptions (see \figref{fig:teaser}), creating a challenging resource for training and evaluating models on detailed video-text compositional alignment. An overview of our data creation pipeline is presented in \figref{fig:overview}.

\subsubsection{Positive video-text pair creation}
\label{sec:positive}
To construct temporally coherent positive video-text pairs, we first sort event captions from ActivityNet-Captions~\cite{krishna2017dense} and YouCook2~\cite{youcook} chronologically by each event’s start time. 
ActivityNet-Captions includes some global descriptions that span multiple events. We filter out captions that cover more than two events. For captions with significant temporal overlap, we compute their temporal IoU. If the IoU exceeds a set threshold, we retain the caption that covers a longer portion of the video.
Since YouCook2 provides temporally distinct, non-overlapping captions, no filtering is necessary. The remaining captions are then concatenated into a single, coherent paragraph that preserves the temporal sequence of events.
To enhance the temporal flow and readability within these paragraphs, we use a large language model (LLM) to add subtle temporal cues such as adpositions, adverbs, or conjunctions. This `LLM-temporal structuring' improves readability without introducing new information or altering the original content.

\subsubsection{Compositional negative sample generation}
\label{sec:negative}
To evaluate compositional understanding, we create diverse negative samples by applying specific disruptions to the temporal and semantic structure positive video-text pairs (see \figref{fig:teaser}). Each disruption alters particular aspects of sequence coherence. 

1) \textbf{Temporal reordering (temp-reorder)} disrupts the chronological order by randomly shuffling event captions within a sequence. We then apply LLM-temporal structuring to maintain linguistic coherence after reordering. 
2) \textbf{Action word replacement (action-replace)} modifies the semantic meaning of the video by selecting a key action or event word and replacing it with a contextually plausible alternative.
3) \textbf{Segment-level mismatch (seg-mismatch)} generates partial sequences by sampling two distinct sub-intervals from the original caption sequence (\eg from [caption-1, 2, 3, 4], we sample intervals [caption-1, 2, 3] and [caption-3, 4]). To ensure each interval retains unique information, we verify that at least two event captions differ between them. We then crop the video sequence to match each text interval, creating distinct video-text pairs. For compositional disruption, we generate negative pairs by mismatching these cropped pairs, pairing the video from one interval with the text from another.
For evaluation, we use the three disruption types above. For training, we also include multi-disruptions, which combine two or more disruption types within a single video-text pair.

As shown in \figref{fig:overview}, we generate negative samples using Gemini-1.5-Pro~\cite{team2024gemini} with carefully designed prompts. These prompts are designed to apply only the specified disruption while preserving the meaning of individual sentences.
For example, the prompt for temporal reordering instructs:
\textit{``Randomly reorder the sentences in this paragraph and add connecting words, such temporal adverbs or transitions, to improve flow. Use only transitions that indicate forward progression in time. Do not use words that suggest backward movement in time.
Ensure that the meaning of each sentence remains unchanged.
Maintain the original number of sentences.”}

To ensure content integrity, we apply automatic validation to check that each negative sample deviates minimally from the original. We also randomly subsample the validation split to reduce evaluation time.
Our benchmark includes approximately 17,000 and 4,500 annotated (video, positive text, negative text) triplets for ActivityNet-Comp and YouCook2-Comp, respectively. Detailed statistics are provided in \tabref{tab:stat}.

\subsubsection{Automatic validation of LLM output}
\label{sec:validation}
To help the LLM-generated outputs for both positive and negative samples maintain the integrity of the original content, we implement an automated validation process that restricts modifications to a minimal threshold. 
For each LLM output, we compare it with the original captions using word-level text comparison.
We use word-level precision and recall between the LM-generated paragraph (P) and the original paragraph (O) as: precision = len(set(P.split()) \& set(O.split())) / len(set(P.split())), recall = len(set(P.split()) \& set(O.split())) / len(set(O.split())).
This closely aligns with the widely-used token F1 metric~\cite{rajpurkar2016squad}.
Outputs falling below 80\% of either precision or recall are not used. 
By applying this process, our benchmark emphasizes compositional coherence without introducing unintended changes.

\begin{table}[t]
\centering
\tablestyle{5pt}{1.2}
 {\fontsize{8pt}{8pt}\selectfont 
\begin{tabular}{lccccc}
\multirow{2}{*}{dataset}           & \multirow{2}{*}{Split} & Temp & Action & Seg & \multirow{2}{*}{Total} \\
    &   & reorder & replace & mismatch  & \\
\toprule
\multirow{2}{*}{ActivityNet-Comp} 
& train    & 5995    & 5583  & 4502 & 16080 \\

 & val     & 280     & 221   & 334   & 835 \\

\multirow{2}{*}{YouCook2-Comp} 
    & train
    & 1174  & 1100  & 1082  & 3356 \\
    & val 
    & 409   & 361   & 397   & 1167 \\
\end{tabular}\vspace{-2mm}
}
\caption{{Number of (video, positive text, negative text) triplets in our benchmark datasets ActivityNet-Comp and YouCook2-Comp.
}}\vspace{-5mm}
\label{tab:stat}
\end{table}

\subsection{Compositionality-aware Learning Objectives}
\label{sec:comploss}
Our model utilizes a dual-encoder CLIP structure for its efficiency and scalability with large-scale video-text data. While cross-attention models might capture compositional relationships in a more intricate way, the dual-encoder approach provides greater flexibility by independently encoding both aligned and disrupted video-text pairs. 
To improve the video CLIP model’s compositional understanding, we extend the standard contrastive loss by introducing a hierarchical pairwise preference loss, termed `CompLoss'. This compositionality-aware learning objective encourages the model to prioritize positive video-text pairs over disrupted ones, assigning progressively lower similarity scores as the disruption level increases.
Our framework starts with the InfoNCE contrastive loss, which serves as the base learning objective. Given a video-text pair (V, T), video and text embeddings are computed independently by the video and text encoders, respectively. These embeddings are then compared within a batch using dot products, scaled by a learnable temperature parameter $\tau$. The video-to-text (V2T) contrastive loss is formulated as follows:
\begin{equation}\label{eqn:contrastive}
L_{\text{V2T}} = -{1 \over {B}} \sum_{i=1}^{B} \log({\text{exp}(V_{i} T_{i} / \tau) \over { \sum_{j=1}^{B} \text{exp}(V_{i} T_{j} / \tau)  }}).
\end{equation}
The text-to-video (T2V) loss mirrors this structure, by swapping the video and text the summation. The total contrastive loss $L_{con}$ is obtained by averaging $L_{con} = (L_{\text{V2T}} + L_{\text{T2V}}) / 2$. 

While approaches like NegCLIP~\cite{yuksekgonul2022and} have explored negative-augmented contrastive losses for image-text compositionality by expanding the text corpus with additional negative texts {T $\cap$ T$_{\text{neg}}$ } in the InfoNCE loss, our method introduces a hierarchical pairwise preference loss tailored to enhance compositional coherence in video-text alignment.

The hierarchical pairwise preference loss, is designed to enforce a similarity ranking across video-text pairs based on their compositional coherence. 
For a positive video-text pair (V, T) and a set of N disrupted negative pairs, denoted as (V, T${\text{neg}}^1$), (V, T${\text{neg}}^2$), ..., where each index denotes the level of disruption applied to the text, the loss establishes a structured hierarchy.
Positive pairs are assigned the highest similarity scores, followed by less disrupted pairs, with heavily disrupted pairs receiving the lowest scores. This hierarchy aligns with the disruption types introduced in \secref{sec:data}, ranging from basic temporal reordering to complex mixed disruptions. The loss $L_{pref}$ is defined as:
\vspace{-2mm}
\begin{equation}\label{eqn:preference}
\begin{split}
L_{pref} = \sum_{i=1}^{N} \max\left(\text{sim}(V, T_{\text{neg}}^i) - \text{sim}(V, T), 0\right) \\
+ \sum_{j > i} \max\left(\text{sim}(V, T_{\text{neg}}^{j}) - \text{sim}(V, T_{\text{neg}}^i), 0\right)
\end{split}
\vspace{-3mm}
\end{equation}
Each T$_{\text{neg}}^i$ represents a disrupted version of the text with increasing degree. The model is trained to encode this hierarchy into the cosine similarity \text{sim}(V, T) between the video V and text T. The highest scores is assigned to positive pairs (V, T), followed by pairs with simple disruptions (\eg temporal reordering), and lowest scores for more heavily disrupted pairs (\eg combinations of reordering and action replacements). This hierarchy helps the model recognize compositional coherence at varying levels of disruption.

The final training objective combines the contrastive loss and the hierarchical pairwise preference loss, balanced by a scaling parameter: $L_{total} = L_{con} + \lambda L_{pref}$. 
This hierarchical preference constraint enables the model to assign similarity scores that reflect the compositional coherence of video-text pairs, enhancing its ability to detect subtle temporal and semantic misalignments across a range of disruptions. By combining contrastive and hierarchical preference objectives, our framework not only captures general video-text associations but also scales alignment strength according to compositional coherence. This approach is especially effective for tasks requiring sensitivity to fine-grained misalignments, supporting a deeper understanding of video-text compositionality.

\subsection{Pretraining with Short-form Video-text Data}
To address the scarcity of densely annotated video-text datasets with temporally localized, multi-event captions, we propose `CompPretrain', a pretraining strategy that utilizes short-captioned video datasets like VideoCC3M~\cite{nagrani2022learning}. These short-form datasets are relatively easier to obtain at scale, and provide an efficient way to approximate the complexity of densely annotated video sequences that are essential for training models on compositional understanding.

\begin{figure}[t]
   \centering
   \includegraphics[width=\linewidth]{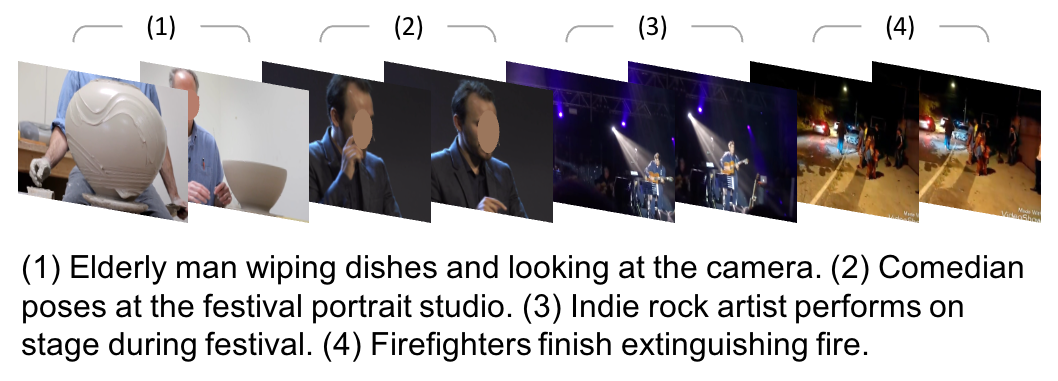}
   \vspace{-6mm}
   \caption{CompPretrain strategy simulates long-form video-text sequences by concatenating short video-text pairs (1)-(4). This enables the use of temporally disrupted sequences and composition-aware learning in video pretraining with the simulated data.}
   \label{fig:comppretrain}
   \vspace{-5mm}
\end{figure}

In this approach, we simulate long-form video-text sequences by stacking multiple short video clips and their captions, creating pseudo long-form structures that mimics the structure of densely annotated datasets (see \figref{fig:comppretrain}). For example, consider three short video-text pairs ($V_1$, $T_1$), ($V_2$, $T_2$), ($V_3$, $T_3$) where each $V_i$ is a short video clip and $T_i$ its corresponding caption. These pairs are combined into a stacked sequence ($V_{\text{stack}}, T_{\text{stack}}$) = ([$V_1$, $V_2$, $V_3$], [$T_1$, $T_2$, $T_3$]) where $T_{\text{stack}}$ describes a multi-event sequence that resembles a continuous temporally organized text.
Within these stacked sequences, we apply compositionality aware techniques from \secref{sec:data} and \secref{sec:comploss} to create negative samples and train the model on compositional coherence. By using the boundaries of each original video segment in the stack, we generate disrupted negative sequences, including temporally reordered sequences where segments within $T_{\text{stack}}$ are randomly shuffled, \eg [$T_2$, $T_1$, $T_3$]. We also create partial captions by removing one or more segments, \eg [$T_1$, $T_2$], that only partially match the video content. These disrupted negative texts are contrasted against the complete $T_{\text{stack}}$ which serves as the positive paragraph.

During pretraining, we apply the pairwise preference loss (\eqnref{eqn:preference}) to encourage the model to assign higher similarity scores to coherent sequences and lower scores to disrupted ones. This enhances the model’s understanding of temporal compositionality, equipping it for tasks requiring fine-grained temporal and semantic alignment.

\subsection{Compositionality Evaluation}
\label{sec:eval}

\paragraph{Evaluation metrics.}\quad
We use two main metrics to evaluate compositionality: video-text retrieval accuracy and binary classification accuracy.
Video-text retrieval accuracy (Recall@1) is a standard measure of general video-text alignment.
For this, we use only the (video, positive text) pairs. This metric reflects how well a model retrieves the correct video for a given text query. It serves as a baseline for assessing overall alignment quality.
For a direct evaluation of compositional understanding, we use binary classification accuracy, adapted from prior work in image-text compositional reasoning~\cite{yuksekgonul2022and} to the video-text setting.
In this setup, each input is a triplet of (video, positive text, negative text). Then, The model must identify which caption better aligns with the video. 
A prediction is correct if the model assigns a higher similarity score to the positive pair. For video CLIP models, we use cosine similarity between video and text embeddings for this comparison.
This binary classification is conducted separately for each disruption type: temporal reordering, action word replacement, and segment-level mismatch. This allows us to assess the model’s robustness across different types of compositional challenges.

In addition to per-disruption accuracy, we report a comprehensive accuracy score across all disruption types. This metric is stricter than simply averaging the binary classification scores. It requires the model to make correct predictions on all types of disruptions. The final score is computed as the product of binary accuracies across all disruption types. This multiplicative score gives a more rigorous measure of compositional robustness, with a random baseline accuracy of 1/2$^N$ for N types of disruptions.

\paragraph{Evaluation of Large Multimodal Models (LMMs).} \quad
To evaluate Large Multimodal Models (LMMs), we adapt the binary classification framework to accommodate their generative output format. For each input, the LMM is presented with a video and its positive and a disrupted caption, and it must select the caption that best aligns with the video content. Binary classification accuracy is then computed for each disruption type, and we calculate a comprehensive accuracy score across all disruption types, following the same multiplicative approach as described above.

\section{Experiments}
\label{sec:experiments}
This section evaluates our model through retrieval on standard benchmarks, compositional alignment on our benchmark, ablations on training and pretraining strategies, and comparisons with large multimodal models.

\subsection{Preliminaries: Baseline Video CLIP model}
\label{sec:baseline}
To build a robust video-text alignment model, we adopt a dual-encoder CLIP architecture for its simplicity and computational efficiency over cross-attention models. We refer to our baseline model as VidCLIP-base. This model starts with the image CLIP architecture pretrained on the DataComp-1B dataset~\cite{gadre2023datacomp}, consisting of a ViT-Large vision encoder (303M) and a 12-layer Transformer text encoder (128M). 
Similar to \cite{coca,videococa}, the vision encoder uses an attention pooler with two layers: the first employs 196 queries, and the second aggregates them into a single global image embedding. The text encoder uses a CLS token to summarize the text features into a global text embedding.
We use an Adam optimizer with a batch size 16k, learning rate of 1e-3, weight decay 1e-2, 224x224 images and train for 500k iterations with linear warmup for 10k iterations.

To adapt the model for video-text alignment, we make several modifications. In the vision encoder, we incorporate temporal positional encoding after the final ViT layer to capture video-level temporal dynamics. 
To generate video representations, we concatenate all frame-level tokens along the temporal dimension into $\mathbb{R}^{B\times (T\times N)\times D}$ which is then processed by the attention pooler to generate a global video embedding.
In the text encoder, we extend the token sequence length from 64 to 160 to support longer video descriptions, also adjusting the sinusoidal positional encoding to match. 
Following these adjustments, VidCLIP-base is further pretrained on the VideoCC3M dataset, which consists of short video-text pairs. For this pretraining, we use an Adam optimizer with a batch size of 128, a learning rate of 1e-7, weight decay of 1e-6, and 50,000 iterations, with 256x256 frame resolution and 16 frames per video.

After pretraining, we evaluate the model’s zero-shot video-text retrieval performance on two widely used benchmarks, MSR-VTT and ActivityNet, comparing it primarily with other dual-encoder models. As shown below, VidCLIP-base achieves competitive results with recent video CLIP models, establishing a strong baseline for further exploration in video-text compositional alignment.

\begin{center}
\tablestyle{10pt}{1.0}
\begin{tabular}{lcc}
Zero-shot  & MSR-VTT    & ActivityNet        \\
Recall@1   & T2V / V2T  & T2V / V2T        \\
\midrule
VideoCLIP~\cite{xu2021videoclip}   & 10.4 / -          & - / -         \\
CLIP4Clip~\cite{luo2022clip4clip}   & 32.0 / -          & - / -         \\
VideoCoCa~\cite{yan2022videococa}   & 34.3 / 64.7       & 34.5 / 33.0   \\
InternVid~\cite{wang2023internvid}     & 42.4 / 41.3       & 32.1 / 31.3   \\

VidCLIP-base (ours)    &  41.8 / 67.0 	   &  30.2 / 30.3  \\
\vspace{-6mm}
\end{tabular}
\end{center}

\begin{table*}[t]
\centering
\small\tablestyle{7pt}{1.0}
\begin{tabular}{llcccc}
{dataset} & {method}   & {Temp-reorder} & {Action-replace} & {Seg-mismatch}   & {All (comprehensive)} \\
\toprule
ActivityNet-Comp 
& VidCLIP-base (zero-shot) & 52.0      & 62.1      & 58.4      & 18.9  \\
& Finetuned	               & 56.6	   & 65.6      & 63.1      & 23.4  \\
& CompLoss        	       & 65.4	   & 73.1      & 65.3      & 31.2  \\
& CompPretrain + CompLoss  & 68.2      & 75.4      & 68.0      & 35.0  \\

\midrule

YouCook2-Comp 
& VidCLIP-base (zero-shot)  & 50.5      & 62.8      & 67.4     & 21.4 \\
& Finetuned	                & 54.2      & 63.9      & 67.0     & 23.2 \\
& CompLoss       	        & 56.3      & 68.8      & 68.5     & 26.5 \\
& CompPretrain + CompLoss   & 58.2      & 70.3      & 69.5     & 28.4 \\

\midrule

\multicolumn{2}{c}{Random guessing} & 50.0 & 50.0 & 50.0 & 12.5 \\

\end{tabular}\vspace{-1mm}
\caption{\small{Evaluation of compositional understanding with our methods CompLoss and CompPretrain, on ActivityNet-Comp and YouCook2-Comp benchmarks. We report binary classification accuracy (\%). Note a random prediction results in a baseline score of 50.0\%.}}
\vspace{-4mm}
\label{tab:tab2}
\end{table*}

\begin{table}[t]
\centering
\small\tablestyle{3pt}{1.0}
\begin{tabular}{lcc}
{method}  & {text-to-video} & {video-to-text} \\
\toprule
VidCLIP-base (zero-shot)     &   27.1 & 30.7	\\
Finetuned	                 &   43.2 & 43.3	\\
CompLoss	                 &   42.4 & 43.0	\\
CompPretrain + CompLoss	    &   42.8 & 43.1	 \\
\end{tabular}\vspace{-1mm}
\caption{\small{Video-text retrieval with the our methods CompLoss and CompPretrain. Recall@1 is reported on ActivityNet-Comp.}}
\vspace{-4mm}
\label{tab:tab3}
\end{table}

\subsection{Evaluation on Compositionality Benchmark}
\vspace{-1mm}
\label{sec:compositionality}
To assess compositional understanding, we conduct binary classification tasks on the ActivityNet-Comp dataset. This evaluation includes the zero-shot performance of our baseline model, VidCLIP-base, as well as the effects of our compositionality-aware techniques, CompLoss and CompPretrain, when fine-tuned on ActivityNet-Comp.
For the fine-tuning process, we use an Adam optimizer with batch size of 32, a learning rate of 1e-6, weight decay of 1e-5, and 20,000 iterations, with a 256x256 resolution and 16 frames per video. When incorporating CompLoss, we set the weighting coefficient $\lambda$ = 100. CompPretrain follows the same pretraining protocol and parameters used for video pretraining on VideoCC3M~\cite{nagrani2022learning}.
Each model variant is evaluated across different types of compositional disruptions: temporal reordering, action replacement, and segment mismatch. We also report a comprehensive score, computed as the multiplicative combination of binary accuracy scores for all disruption types, as described in \secref{sec:eval}.

As shown in \tabref{tab:tab2}, each improvement results in gradual gains, with CompLoss and CompPretrain yielding clear benefits across all disruption types. Our proposed  datasets ActivityNet-Comp and YouCook2-Comp, when combined with these techniques, effectively enhance the model's robustness against compositional disruptions.

We further evaluate video-text retrieval on ActivityNet-Comp. We use only positive pairs with the LLM-temporal structured captions. \tabref{tab:tab3} presents Recall@1 for each method.
Standard finetuning achieves the highest scores, likely because it is trained only on the contrastive objective. CompLoss and CompPretrain show similar retrieval performance, suggesting that compositional training does not cause a noticeable drop in retrieval accuracy.

\begin{figure}[t]
   \centering
   \includegraphics[width=\linewidth]{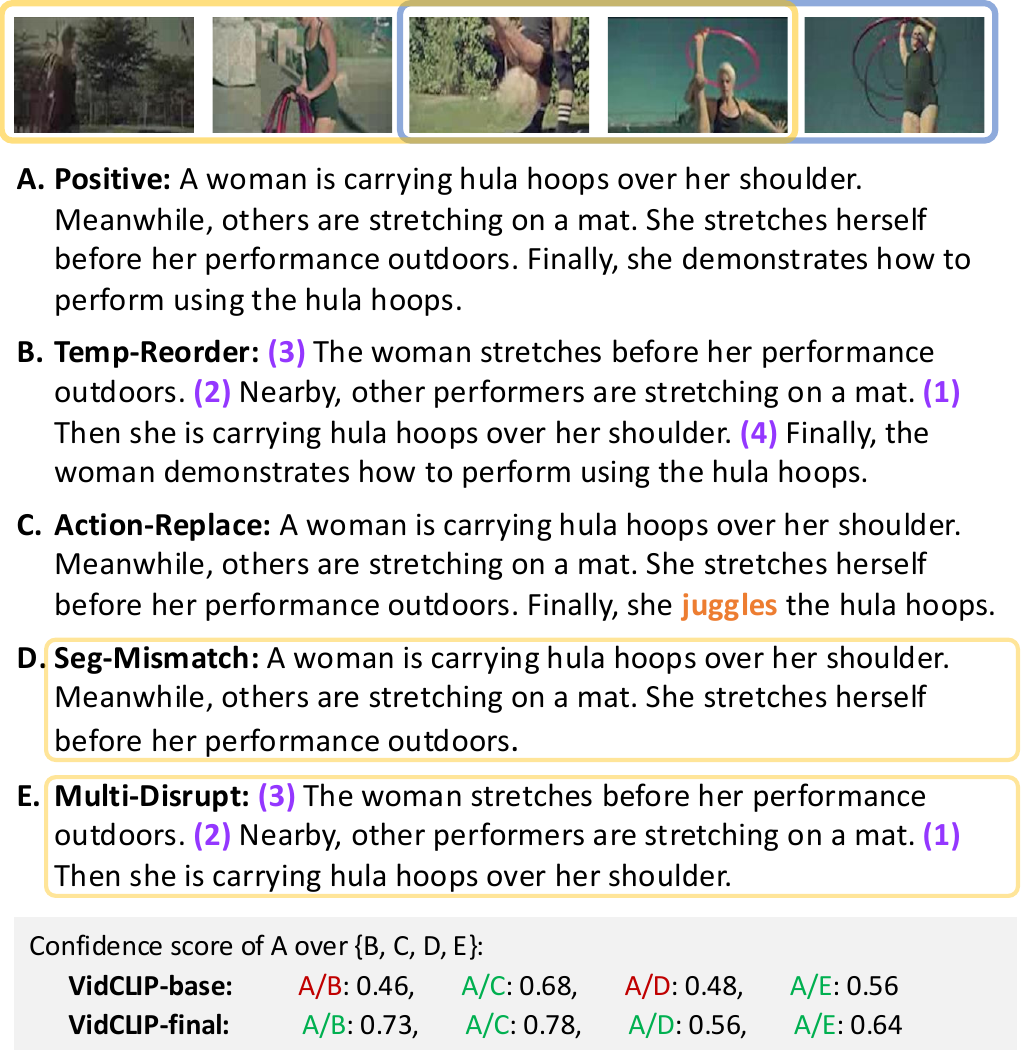}
   \vspace{-6mm}
   \caption{\small{Comparison of VidCLIP-base and VidCLIP-final on our benchmark dataset, with their confidence scores in selecting the positive text (A) over various negative samples (B, C, D, E). Temporal reordering (purple), action replacement (orange), segment mismatch (text in yellow box matched with video crops in blue), and multiple disruptions. VidCLIP-final consistently achieves higher scores, across different compositional disruptions.}}
   \label{fig:result}
   \vspace{-2mm}
\end{figure}

\figref{fig:result} illustrates the results of VidCLIP on our benchmark dataset, comparing the baseline (VidCLIP-base) and the full method (VidCLIP-final) which is pretrained with CompPretrain, then finetuned on our Comp dataset using CompLoss. The figure shows confidence scores for selecting the positive text A over various negative texts B, C, D, E: temporal reordering, action replacement, segment cropping, and multiple disruptions, respectively. VidCLIP-final  achieving higher confidence scores across all types of compositional disruptions, showing improved robustness in video-text alignment.

\subsection{Ablation Studies}
We conduct ablation studies to evaluate the effectiveness of each compositionality-focused component, CompLoss and CompPretrain, on the ActivityNet-Comp dataset. Each study examines how these elements impact the model’s ability to achieve compositional alignment and robustness across disruption types.

\paragraph{Effect of compositionality-aware learning objectives (CompLoss).}\quad
To analyze the effect of CompLoss on compositional alignment, we compare it against the baseline model finetuned with standard contrastive loss, and a negative-augmented contrastive loss from NegCLIP~\cite{yuksekgonul2022and}.
As shown in \tabref{tab:tab4}, CompLoss which includes a hierarchical pairwise preference loss, outperforms both the baseline contrastive loss and NegCLIP across all disruption types.

\begin{table}[t]
\centering
\small\tablestyle{5.0pt}{1.0}
\begin{tabular}{lcccc}
\multirow{2}{*}{method} & {Temp} & {Action} & {Seg} & \multirow{2}{*}{All} \\
& {reorder} & {replace} & {mismatch} & \\
\toprule
Standard contrastive loss   & 52.0      & 62.1      & 58.4      & 18.9  \\
NegCLIP                     & 60.0	   & 72.7      & 62.6      & 27.3  \\
CompLoss                    & 65.4	   & 73.1      & 65.3      & 31.2  \\

\end{tabular}\vspace{-1mm}
\caption{\small{Ablation on CompLoss (compositionality-aware learning). Binary classification accuracy (\%) on ActivityNet-Comp.}}
\vspace{-4mm}
\label{tab:tab4}
\end{table}

\paragraph{Effect of pretraining strategy (CompPretrain).}\quad
To evaluate the influence of CompPretrain on compositional alignment, we conduct experiments in the zero-shot setting after pretraining on the VideoCC3M dataset~\cite{nagrani2022learning}.
\tabref{tab:tab5} shows that CompPretrain significantly improves the model's robustness across all disruption types, both in the zero-shot setting and when fine-tuned with CompLoss. This improvement highlights the effectiveness of pretraining on stacked, short-form video-text sequences, enabling the model to better generalize in capturing fine-grained video-text alignment.
\tabref{tab:tab5} shows that using a stack size of four leads to better compositional alignment scores compared to a stack size of one.

We further evaluate video-text retrieval performance in a zero-shot setting to examine the effectiveness of our pretrained representations in video-text alignment. We evaluate on multiple datasets, including our ActivityNet-Comp and YouCook2-Comp benchmarks with longer multi-event video-text pairs, as well as the short-captioned MSR-VTT dataset. As shown in \tabref{tab:tab6}, CompPretrain is effective across both short and long video-text pairs, yielding significant improvements on longer sequences in ActivityNet-Comp and YouCook2-Comp. It also provides a performance boost on MSR-VTT, demonstrating adaptability to varying sequence lengths and complexities. These suggests that CompPretrain strategy enhances model robustness for complex video-text pairs without sacrificing performance on shorter sequences.

\begin{table}[t]
\centering
\small\tablestyle{3pt}{1.0}
\begin{tabular}{lccccc}
\multirow{2}{*}{method} & {CompPretrain} & {Temp} & {Action} & {Seg} & \multirow{2}{*}{All} \\
& {stack size} & {reorder} & {replace} & {mismatch} & \\
\toprule
\multirow{2}{17mm}{VidCLIP-base (zero-shot)}  
&  1 (None) & 52.0      & 62.1      & 58.4      & 18.9  \\
&  4        & 54.5      & 64.8      & 61.1      & 21.6  \\
\midrule
CompLoss      & 1 (None)    & 65.4	   & 73.1      & 65.3      & 31.2  \\
(finetuned)   & 4           & 68.2     & 75.4      & 68.0      & 35.0  \\
\end{tabular}\vspace{-2mm}
\caption{\small{Ablation on CompPretrain strategy. Binary classification accuracy (\%) on ActivityNet-Comp.}}
\vspace{-1mm}
\label{tab:tab5}
\end{table}

\begin{table}[t]
\centering
\tablestyle{10pt}{1.0}
\begin{tabular}{lcc}
\multirow{2}{*}{method}   & MSR-VTT   & ActivityNet-Comp  \\
{}   & T2V / V2T & T2V / V2T   \\
\midrule
VidCLIP-base    & 41.8 / 67.0   & 27.1 / 30.7 \\
CompPretrain    & 43.6 / 68.0   & 31.9 / 35.0 \\
\end{tabular}
\vspace{-2mm}
\caption{\small{Effect of CompPretrain on zero-shot video-text retrieval. Recall@1 is reported.}}
\vspace{-4mm}
\label{tab:tab6}
\end{table}

\subsection{Comparison with Large Multimodal Models}
\vspace{-1mm}
We evaluate several video-text foundation models and large multimodal models (LMMs) on our benchmark in a zero-shot setting. \tabref{tab:tab7} presents binary classification accuracy on ActivityNet-Comp across temporal reordering, action word replacement, and segment-level mismatch, along with the comprehensive score.

For LLM evaluation, we use a prompt such as:
{{\textit{Given the video and two text candidates, your task is to determine which paragraph better aligns or matches with the video content.
Candidate 1: \{paragraph\_1\}
Candidate 2: \{paragraph\_2\} 
Which text candidate better matches the video content? Answer with "1" or "2" only.}}}
Here, {\textit{paragraph\_1}} and {\textit{paragraph\_2}} are randomly shuffled positive and negative paragraphs. The model is evaluated based on whether it outputs the correct answer digit (``1" or ``2") as an exact match.

\begin{table}[t]
\centering
\small\tablestyle{4pt}{1.00}
\begin{tabular}{lcccc}
\multirow{2}{*}{method and \# params}   & {Temp} & {Action} & {Seg} & \multirow{2}{*}{All} \\
{}   & {reorder} & {replace} & {mismatch} & {} \\
\toprule
\textit{Zero-shot:} & & & & \\
VidCLIP-base (ours) $\sim$500M     & 52.0      & 62.1      & 58.4      & 18.9  \\
CLIP4Clip~\cite{luo2022clip4clip} $\sim$200M    & 51.2 & 60.2 & 55.3 & 17.0 \\
VideoCoCa~\cite{yan2022videococa} $\sim$2B  & 53.1  &  64.3  &  58.9  & 20.1 \\
VideoPrism~\cite{zhao2024videoprism}       & 53.4   & 66.9  & 62.2 & 22.2 \\

Gemini-1.5-Flash-8B~\cite{team2024gemini}  & 67.1 & 79.6 & 67.3 & 35.9 \\
Gemini-1.5-Flash~\cite{team2024gemini}  & 69.6 & 83.3 & 73.3 & 42.5 \\
Gemini-1.5-Pro~\cite{team2024gemini}    & 70.4 & 84.2 & 74.1 & 43.9 \\
\midrule
\textit{Finetuned on our Comp dataset:} & & & & \\
VideoCLIP-final (ours) $\sim$500M	    & 68.2      & 75.4      & 68.0      & 35.0  \\
\bottomrule
\end{tabular}\vspace{-2mm}
\caption{\small{Comparison of compositional video-text alignment on ActivityNet-Comp. Each input video consists of 16 uniformly sampled frames. Binary classification accuracy (\%) is reported.}}
\vspace{-4mm}
\label{tab:tab7}
\end{table}

Our compositionality benchmark remains highly challenging, even with lightweight negative sample generation strategies. 
With a 50\% random baseline in binary classification, strong foundation models like VideoCoCa~\cite{videococa} and VideoPrism~\cite{zhao2024videoprism} achieve only mid-50s to 60s accuracy.
Even large LMMs reach only the 60s to low 70s on tasks like temporal reordering and segment mismatch, highlighting the difficulty of achieving fine-grained temporal alignment. 
We also observe that larger LMMs perform better overall: Gemini-1.5-Pro $>$ Flash $>$ Flash-8B. This suggests that our benchmark effectively reflects model capacity and compositional reasoning capabilities.

Compared to our finetuned VidCLIP-final model, LMMs perform better on action word replacement, showing their strong grasp of object- and action-level semantics. However, the 8B model lags behind on temporally sensitive tasks like temporal reordering and segment mismatch, where our model shows improved robustness through compositionality-aware training.
Overall, these results show that our benchmark effectively reveals limitations in current video-text models and highlight the strength of our approach in capturing compositional alignment.

\section{Conclusion}
\label{sec:conclusion}

We introduce a benchmark and training framework to improve video-text compositionality, focusing on temporal coherence and alignment. Using dense video captioning datasets, we create ActivityNet-Comp and YouCook2-Comp with compositional disruptions to test models' ability to detect misalignments. We also propose learning methods, CompLoss and CompPretrain, to improve sensitivity to temporal and semantic inconsistencies. Comparisons with large multimodal models highlight challenges and demonstrate the effectiveness of our approach.

{
    \small
    \bibliographystyle{ieeenat_fullname}
    \bibliography{main}
}

\end{document}